\documentclass{article}
\usepackage{rotating}
\usepackage{algorithmic}
\usepackage[ruled]{algorithm2e} 
\usepackage{float}

\usepackage[caption = false]{subfig}
\usepackage{graphicx}
\usepackage{flushend}
\begin{document}

\title{\large\bf A Novel Spatial-Spectral Framework for the Classification of Hyperspectral Satellite Imagery}
\author{Shriya TP Gupta\footnote{BITS, Pilani, Dept. of CS \& IS, Goa Campus, Goa, India, Email: shriyatp99@gmail.com} \hspace{0.15mm} and Sanjay K. Sahay\footnote{BITS, Pilani, Dept. of CS \& IS, Goa Campus, Goa, India, India, Email: ssahay@goa.bits-pilani.ac.in}}

\date{}

\maketitle              
\begin{abstract}
Hyperspectral satellite imagery is now widely being used for accurate disaster prediction and terrain feature classification. However, in such classification tasks, most of the present approaches use only the spectral information contained in the images. Therefore, in this paper, we present a novel framework that takes into account both the spectral and spatial information contained in the data for land cover classification. For this purpose, we use the Gaussian Maximum Likelihood (GML) and Convolutional Neural Network (CNN) methods for the pixel-wise spectral classification and then, using segmentation maps generated by the Watershed algorithm, we incorporate the spatial contextual information into our model with a modified majority vote technique. The experimental analyses on two benchmark datasets demonstrate that our proposed methodology performs better than the earlier approaches by achieving an accuracy of 99.52\% and 98.31\% on the Pavia University and the Indian Pines datasets respectively. Additionally, our GML based approach, a non-deep learning algorithm, shows comparable performance to the state-of-the-art deep learning techniques, which indicates the importance of the proposed approach for performing a computationally efficient classification of hyperspectral imagery.
\vspace*{0.1cm}
~
\end{abstract}

\section{Introduction}

Hyperspectral satellite imaging collects high resolution images within contiguous spectral bands across a wide range of the electromagnetic spectrum. Its primary aim is to obtain the spectrum of each pixel in the image of a scene, with the goal of identifying materials or discovering objects. In general, hyperspectral satellite images are captured through an imaging spectrometer in narrow bands of 10--20 nm with up to thousands of bands. On the contrary, multispectral images have 3--10 bands and are captured using a remote sensing radiometer. Such remotely sensed satellite images, either air-borne or space-borne, cover large areas of the earth's surface with rich spectral information and hence can be used for the robust classification of land cover with a good accuracy.

Recently, machine learning and deep learning techniques have shown promising results for various classification problems in the fields of natural language processing, computer vision, speech recognition, etc.~\cite{shrestha2019review}. Thus, we investigate both, GML, a machine learning technique as well as CNN, a popular deep learning approach. In conventional pixel-wise classification techniques, the algorithms independently process each and every pixel without using the neighboring spatial information. However, to minimize the uncertainty in the data, one can exploit the spatially adjacent pixels, which have more or less similar spectral features. Therefore, in this paper, we develop a novel spatial-spectral algorithm for the robust classification of land cover. Experimental results show that our model is suitable for the classification of images with large spatial structures especially when the spectral responses of different classes are dissimilar. Hence, our proposed approach works well for the task of hyperspectral image classification and it can also be adapted to other domains that are similar to terrain feature classification.

The rest of this paper is organized as follows: Section 2 covers the related work and in Section 3, we give a brief outline of the classification and segmentation methods. Section 4 describes the datasets used and the detail of our approach is given in Section 5. In Section 6, we discuss our experimental results and finally, Section 7 contains the conclusion and future directions.

\section{Related Work}

Spectral imaging techniques are now widely being used for solving land cover classification problems. In this context, Ahmad et al. \cite{ahmad2012analysis} have shown that if there exists a distinct separation between the classes in the decision space, as seen in the case of land cover classification tasks, the use of a GML classifier provides a good accuracy. Further, Ablin et al. \cite{ablin2013survey} reported that classifying an unknown pixel can be done efficiently by the maximum likelihood classifier, as it quantitatively evaluates both the variance and co-variance of the spectral response pattern for each class.

In 2018, Khan et al. \cite{khan2018modern} discussed in detail the recent research in classification techniques used for hyperspectral satellite imagery. In their paper, they state that most of the current research efforts and studies follow the standard pattern recognition paradigm, which is based on the construction of complex handcrafted features. In contrast to the approaches discussed by Khan et al. \cite{khan2018modern}, LeCun et al. \cite{lecun2015deep} proposed the deep learning based classification methods which use hierarchically constructed high-level features in an automated way. In the context of deep learning based satellite image classification, Makantasis et al. \cite{makantasis2015deep} used a combination of CNN for feature extraction along with a multi layer perceptron for classification and achieved an accuracy of up to 98.88\% on the Indian Pines dataset. Similarly, Audebert et al. \cite{audebert2019deep} used 3D CNNs for the classification of Indian Pines and Pavia University datasets, and obtained accuracies of 96.87\% and 96.71\% respectively. An alternate approach of using Recurrent Neural Networks (RNN) was proposed by Ma et al. \cite{ma2019hyperspectral}. Their RNN based model achieved an accuracy of 96.20\% on the Pavia University dataset which indicates the superiority of CNNs over RNNs for image classification tasks. 

Later, many authors combined the aforementioned pixel-wise classification algorithms with different segmentation approaches to enhance the efficiency of the spectral classification. For instance, Tarabalka et al. \cite{tarabalka2009spectral} studied the use of Watershed segmentation methodologies for spatial-spectral classification of hyperspectral images. They also described the various plausible approaches to combine it with a pixel-wise classifier, viz. they used a SVM classifier with Watershed segmentation and reported an accuracy of 90.64\% on the Indian Pines image and 95.21\% on the University of Pavia dataset. Furthermore, some authors have explored the use of Markov Random Fields (MRF) to incorporate the spatial information contained in the data. For example, Qing et al. \cite{qing2018spatial} used an MRF-based loopy belief propagation technique and obtained an accuracy of 98.5\% on the Indian Pines dataset.

Hence, we consider these approaches as the baselines for our work and in our proposed model, we further improve the classification accuracy as well as the computational efficiency using both, GML, a machine learning technique as well as CNN, a deep learning approach for pixel wise classification. To the best of our knowledge, land cover classification has rarely been done by using the watershed segmentation algorithm with the above-mentioned spectral classifiers.

\section{Preliminaries}

In this section, we briefly explain the GML and CNN models used for the pixel-wise classification as well as the Watershed algorithm used to generate the segmentation maps.

\subsection{Gaussian Maximum Likelihood}
\label{GMLsection}

The Gaussian Maximum Likelihood classifier is a supervised classification method derived from the Naive-Bayes theorem. It classifies each unidentified pixel $\theta$ by using the probability density functions to compute the likelihood of a given pixel belonging to a particular category C. The GML algorithm evaluates the probability values for each class as follows:
\begin{equation}
\label{eq:bayes}
P(C|\theta) = P(C ) \frac{P(\theta|C)}{P(\theta)}
\end{equation}
 Here, the pixel is assigned to the most likely class based on the highest probability value or it is labelled as `unknown' if the probability values are below an analytically defined threshold. As the GML is a pixel by pixel classification method, it does not take into account the contextual information about neighboring classes while labeling a pixel. Also, it is assumed that the distribution of the cloud of points forming a particular class is Gaussian. Here, the mean vector and the co-variance matrix can be used to entirely describe the distribution of the response pattern for each category.

\subsection{Convolutional Neural Network}
\label{CNNsection}

A Convolutional Neural Network is comparatively more accurate than a traditional machine learning algorithm mainly because a CNN learns the filters automatically while the features of a traditional algorithm are hand-engineered. Here, we briefly describe the working of a CNN with hyperspectral image data as the input and the detailed description of its usage in our model is presented in Section~\ref{Methodology}. Since a hyperspectral image is composed of $L$ spectral layers, the input image can be represented as X = $[x_1, x_2, . . . , x_L ]$ where $x_i$ is a matrix of pixels corresponding to the layer i. The first convolutional layer consists of $n_1$ filters of kernel size $f_1\times f_1$ and it is employed to extract the features of the input image as follows:
\begin{equation}
\label{eq:f1}
F_1(X) = \theta_1(u_1)  
\end{equation}
 where  $\theta_1$ denotes the activation function of the entire layer. Here $u_1 = \sum_{l=1}^{L} x_l * w_l + b_1$, where $w_l$ represents the weights of the filter acting on the input hyperspectral data, $*$ represents the convolution operation, and $b_1$ is an $n_1$-dimensional vector which represents the bias of the entire layer. Also, each element of this vector is associated with every filter and the output is composed of $n_1$ feature maps corresponding to the $n_1$ filters.

\subsection{Watershed transformation}
\label{Watershedsection}

The Watershed transformation is a powerful mathematical technique for geomorphological image segmentation. It considers a 2D one-band image as a topographic structure wherein the value of a pixel stands for its elevation. In order to associate each basin with a minima, the Watershed lines divide the entire image into catchment basins. This segmentation method requires the user to define seeds, which are pixels belonging unambiguously to a region. The Watershed transformation is usually applied to the gradient function of the image. The gradient defines the transitions between regions, so that it has high values on the borders between objects and minima in the homogeneous regions. If the crest lines in the gradient image correspond to the edges of image objects, then the Watershed transformation successfully partitions the entire image into meaningful regions. Hence, it is very well suited for hyperspectral satellite image segmentation where the main aim is to segment the imagery into distinct geographical regions. 

\section{Datasets and Pre-processing}

In this section we describe the two publicly available benchmark datasets (Indian Pines~\cite{landgrebe1992aviris} and Pavia University~\cite{pavia}) and the preprocessing that has been used for our experimental analysis.

\subsection{Indian Pines}

The Indian Pines dataset consists of $145\times145$ pixels and 224 spectral reflectance bands and was gathered by the 220-band Airborne Visible / Infrared Imaging Spectrometer (AVIRIS) sensor over the Indian Pines test site in North-western Indiana. The ground truth corresponding to this dataset is designated into sixteen classes which is not entirely mutually exclusive and consists of two-thirds agriculture and one-third forest or other natural perennial vegetation. Along with this, there are two dual lane highways, a rail line, low density housing, and some smaller roads. 

\subsection{Pavia University}

The Pavia University dataset was acquired by the 103-band Reflective Optics System Imaging Spectrometer (ROSIS) sensor during a flight campaign over Pavia, northern Italy with dimensions of $610\times610$ pixels and 103 spectral bands. However, some of the samples contain no information. Hence, they are discarded in our experimental analysis and are depicted using broad black strips. The geometric resolution of this image is 1.3 meters and the corresponding ground truth differentiates among all the nine classes. 

\subsection{Pre-processing}

For the empirical analysis, the preprocessing steps of the Indian Pines and the Pavia dataset are same for both the approaches (GML and CNN). First, we over sample the weak classes in order to make the hyperspectral images more balanced. The pixel values are then standardized by subtracting the mean and scaling them to unit variance. Next, we apply Principal Component Analysis (PCA) to the input data in order to transform the original high-dimensional data into a low-dimensional space by extracting the key features and then padding it with zeros. Then, we save the preprocessed data for the experimental analysis.

\section{Methodology}
\label{Methodology}

This section describes our approach in detail which includes the pixel-wise classification techniques, the segmentation method, and the modified majority vote algorithm.

\begin{figure}[!thpb]
      \centering
      \includegraphics[scale=0.47]{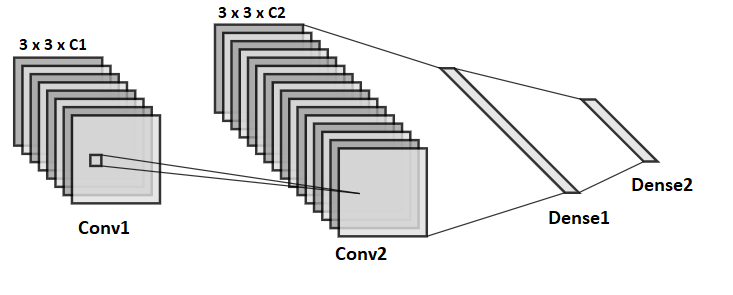}
      \caption{An overall architecture of the CNN }
      \label{CNNnet}
\end{figure}

\subsection{Spatial-Spectral Classifier}

 An overall architecture of the CNN used in our work is shown in Figure~\ref{CNNnet}. It comprises of two 2D convolutional layers with the first layer having $C_1$ filters of size $f_1 = 3$, second layer having $C_2 = 3 \times C_1$ filters of size $f_2 = 3$, and followed by a flatten layer. Here, the parameter $C_1$ corresponds to the number of principle components that preserve at least 99.9\% of the initial information contained in the original datasets after applying PCA. The size of the filter is determined empirically which dictates the number of neighbors of each pixel that has been taken into consideration during the classification. The convolution layer is composed of a set of separate filters and each filter is independently convolved with the image to obtain the corresponding feature maps as described in Section~\ref{CNNsection}. In contrast to the conventional CNNs, we do not use a max pooling layer after the convolution layer, since there is no need to account for the translation and scale invariance. Next, we use two regular densely connected neural network layers in our model. The first dense layer has 120 units with a dropout of 0.5 and the second layer is the last layer of the network. It has as many units as the number of classes and in this layer, we use a softmax algorithm to produce the final outputs.

The second pixel-wise classification scheme that we used is the GML algorithm as outlined in Section~\ref{GMLsection}, and in our analysis, the GML classification technique differs significantly from the CNN method. The GML algorithm performs the classification in a pixel by pixel fashion which preserves their individual values. On the contrary, the convolution operation in the CNN technique averages out individual values using adjacent pixels that lie within the dimensions of an applied filter.

\begin{figure}[!thpb] 
\centering
\subfloat[]{\includegraphics[width=.14\linewidth]{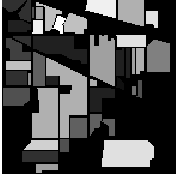}}
\hspace{2em}
\subfloat[]{\includegraphics[width=.14\linewidth]{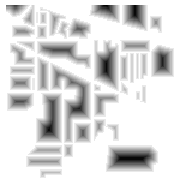}}
\hspace{2em}
\subfloat[]{\includegraphics[width=.14\linewidth]{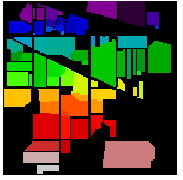}}
\caption{(a)Overlaps (b)Distances and the (c)Separated objects for Indian Pines data}
\label{watershed}
\end{figure}

\begin{figure}[!thpb] 
\centering
\subfloat[]{\includegraphics[width=.13\linewidth]{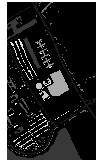}}
\hspace{2.8em}
\subfloat[]{\includegraphics[width=.13\linewidth]{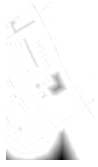}}
\hspace{2.8em}
\subfloat[]{\includegraphics[width=.13\linewidth]{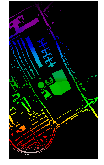}}
\caption{(a)Overlaps (b)Distances and the (c)Separated objects for the Pavia dataset}
\label{watershedpavia}
\end{figure}

Next, we generate a segmentation map of the image using the Watershed algorithm as detailed in Section~\ref{Watershedsection}. In our model, local maxima of the distance map to the background are used as seeds. These maxima are chosen as markers in order to segment the entire image into regions and the flooding of catchment basins from such markers separates the areas along a Watershed line. The segmentation maps obtained using the Watershed algorithm are shown in Figure~\ref{Watershed} and Figure~\ref{Watershedpavia} for the Indian Pines and the Pavia University datasets respectively. For images where the pixels are not uniformly distributed across the classes, the Watershed algorithm typically produces over segmented maps, and so in our case, the image is over segmented. However, the problem is alleviated when we combine the segmentation map with the pixel-wise classification results.

\begin{algorithm}
 \label{algo}
 \caption{Our majority vote approach.}
 \begin{algorithmic}[1]
    \STATE uni\_labels = unique values (labels in segmentation map)
    \STATE final\_map = 0
    \FOR{i in uni\_labels}
    \STATE arr $\leftarrow$ []
    \STATE height $\leftarrow$ height of segmentation map
    \STATE width $\leftarrow$ width of segmentation map
    \FOR{j = 0 to j = height }
    \FOR {k = 0 to k = width }
    \STATE find all pixels in the segmentation map having label i
    \STATE append their pixel-wise classification labels to arr
    \ENDFOR
    \ENDFOR
    \STATE max\_freq\_label $\leftarrow$ most frequent label in arr
    \FOR{j = 0 to j = height}
    \FOR{k = 0 to k = width}
    \STATE find all pixels in the segmentation map having label i
    \STATE assign max\_freq\_label as the label for these pixels in the final\_map
    \ENDFOR
    \ENDFOR
    \ENDFOR
    \RETURN final\_map
 \end{algorithmic} 
\end{algorithm}

Finally, we combined the outputs of the pixel-wise classification algorithms with the segmentation maps using a modified majority vote technique (Algorithm~\ref{algo}), wherein for each region of the segmentation map, all the pixels are assigned to the most frequent class. However, in our proposed approach, the majority vote is performed with an adaptive neighborhood instead of a fixed neighborhood and for every pixel, the region it belongs to is defined by the segmentation map. Then, we use this region as its neighborhood for the vote on the spectral classification.

\section{Experimental Results}

First, we test our proposed models, GML with Watershed (GML-W) and CNN with Watershed (CNN-W), on the Indian Pines and the Pavia University datasets. Then, we compare the performance of our hybrid models with the traditional GML and CNN algorithms on the same datasets. For this purpose, we split the datasets in the ratio of 3:1 for the training and testing data respectively. 

\begin{figure}[!thpb] 
\centering
\subfloat[]{\includegraphics[width=.17\linewidth]{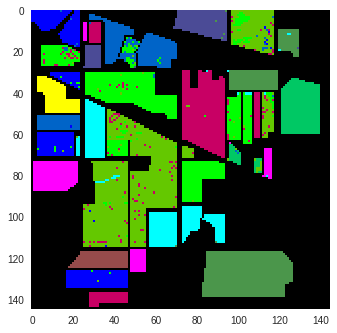}}
\hspace{1.2em}
\subfloat[]{\includegraphics[width=.17\linewidth]{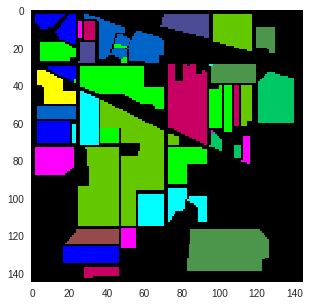}}
\hspace{1.2em}
\subfloat[]{\includegraphics[width=.17\linewidth]{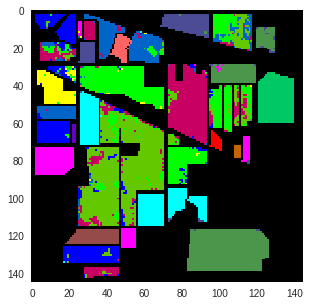}}
\hspace{1.2em}
\subfloat[]{\includegraphics[width=.17\linewidth]{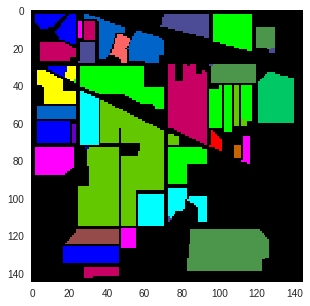}}
\caption{Results of (a)GML (b)GML-W (c)CNN and (d)CNN-W on Indian Pines data}
\label{GML for Aviris dataset}
\end{figure}

The results obtained by the traditional GML and the proposed GML-W algorithms are shown in Figure~\ref{GML for Aviris dataset}(a) and Figure~\ref{GML for Aviris dataset}(b) respectively. For the Indian Pines dataset, the GML gives an accuracy of 97.46\% and the GML-W provides an accuracy of 98.31\%. Next, we experimented with the CNN classifier and the obtained results are shown in Figure~\ref{GML for Aviris dataset}(c) and Figure~\ref{GML for Aviris dataset}(d). In terms of accuracies, the CNN and CNN-W models give results of 87.08\% and 97.7\% respectively. As is evident from the results, the use of the spatial-spectral classifier led to a significant improvement with an increase of 10.6\% in the overall accuracy. Additionally, from a visual observation, the classification maps obtained by the spatial-spectral classification are seen to be much less noisy than the ones obtained by the pixel-wise classification.

\begin{table}[!thpb]
\renewcommand{\arraystretch}{1.0}
\caption{Accuracies (\%) for Indian Pines dataset}
\label{IndianPines}
\centering
 \begin{tabular}{|c|c|c|c|c|} 
 \hline
Category & GML  &  GML-W  & CNN &  CNN-W \\[0.5ex]
\hline
Alfalfa               &    100.0   &   100.0  &    100.0    &  100.0 \\
Corn-notill           &   0.00      &   100.0   &   80.11   &    0.00\\
 Corn-mintill         &   94.74   &   82.35   &   89.90    &   100.0\\
 Corn                 &    97.83  &   88.19   &   96.61   &     93.37\\
Grass-pasture         &    100    &   100.0   &    95.04    &   100.0\\
Grass-trees           &    99.79  &    96.89   &   98.90    &   96.89\\
Grass-pasture-mowed   &    99.72  &   100.0   &  100.0  & 100.0 \\      
Hay-windrowed         &   0.00       &   100.0   &   100.0  &  0.00\\
  Oats                &   100.0     &   100.0   &   100.0   & 100.0  \\
Soybean-notill        &   0.00       &   100.0  &    88.06   &   0.00\\
Soybean-mintill       &   99.48   &   99.48   &   72.14   &    99.48\\
Soybean-clean         &   90.59   &   98.20   &   87.83   &    98.28\\
 Wheat                &   99.32   &   100.0   &   100.0   &     100.0\\
 Woods                &   100.0      &   100.0   &   98.41  &    100.0\\
Buildings-Grass-Trees &    99.60   &   99.20    &  93.81  & 99.84\\
Stone-Steel-Towers    &    98.96   &   87.56    &  100.0   &    87.04\\
\hline
Average accuracy          &  93.80   &   96.99 &     80.10  &    91.76\\
Overall accuracy             &   97.46   &   98.31   &   87.08  &    97.7\\
 \hline
\end{tabular}
\end{table}

The detailed results obtained for each category, viz. GML, CNN, GML-W, and CNN-W on the Indian Pines dataset is shown in Table~\ref{IndianPines}. From the results, it is evident that incorporating the spatial information with the traditional pixel-wise classification algorithm improves the accuracies considerably. This is due to the fact that most of the classes in the image represent large crop fields, and the segmentation step makes these regions of fields homogeneous, thereby improving the classification results. The accuracies of almost all classes are significantly improved, except for some small classes like Oats and Stone-Steel Towers which are incorrectly assimilated into neighboring regions. However, the GML algorithm fails to correctly identify the regions of Corn-notill, Hay-windrowed, and Soybean-notill as the number of pixels in the training set is almost a hundred times lesser than other regions, which leads to the misclassification of the pixels into neighboring categories.

\begin{figure}[!thpb] 
\centering
\subfloat[]{\includegraphics[width=.17\linewidth]{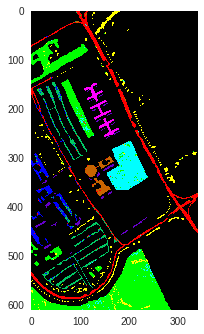}}
\hspace{1.3em}
\subfloat[]{\includegraphics[width=.17\linewidth]{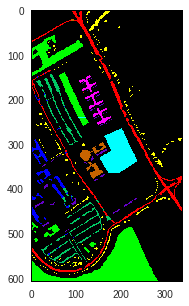}}
\hspace{1.3em}
\subfloat[]{\includegraphics[width=.17\linewidth]{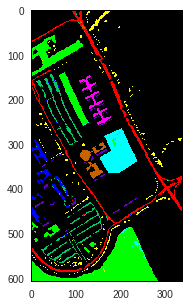}}
\hspace{1.3em}
\subfloat[]{\includegraphics[width=.17\linewidth]{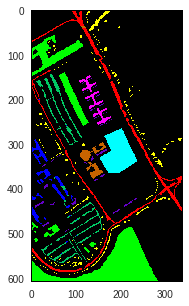}}
\caption{Results of (a)GML (b)GML-W (c)CNN and (d)CNN-W on the Pavia dataset}
\label{GMLC for Pavia dataset}
\end{figure}

A similar analysis of the Pavia University dataset has been done by splitting the data in the ratio of 3:1 for training and testing. First, using the GML classifier, we were able to obtain an accuracy of 99.03\% which was later on improved to 99.52\% using our proposed hybrid classifier and the results are shown in Figure~\ref{GMLC for Pavia dataset}(a) and Figure~\ref{GMLC for Pavia dataset}(b), respectively. Next, we carried out the pixel-wise classification using a deep learning based CNN model and obtained an accuracy of 97.39\%. We further improved the performance by $\sim$ 2\% using our proposed approach which led to an overall accuracy of 99.44\%. Figure~\ref{GMLC for Pavia dataset}(c) and Figure~\ref{GMLC for Pavia dataset}(d) represent the results obtained before and after combining with the Watershed transformation respectively.

\begin{table}[!thpb]
\renewcommand{\arraystretch}{1.0}
\caption{Accuracies (\%) for the Pavia university dataset}
\label{Pavia}
\centering
 \begin{tabular}{|c|c|c|c|c|} 
 \hline
Category & GML  &  GML-W  & CNN &  CNN-W \\[0.5ex]
\hline
 Asphalt             &    100.0   &   100.0   &   98.13   &   100.0\\
Meadows              &    95.02  &    98.02   &   98.62    &  97.49\\
Gravel               &   94.98   &   96.02   &   89.33   &    96.02\\
Trees                &   92.42    &  96.56    &  99.73    &   91.47\\
Painted metal sheets &   99.18  &    99.18  &    100.0    &   99.18\\
 Bare Soil           &    100.0    &  100.0 &     98.96  &    100.0\\
Bitumen              &   94.96   &   100.0   &   99.39   &    100.0\\
Self-Blocking Bricks &    96.76   &   100.0  &    93.81    &   100.0\\
Shadows              &    92.85   &   99.10   &   100.0   &    98.66\\
\hline
Average accuracy         &    96.24 &    98.76   &   97.55  &   98.09\\
Overall accuracy     &    99.03   &   99.52    &  97.39   &  99.44\\
\hline
\end{tabular}
\end{table}

Furthermore, we did a detailed analysis of our models for the Pavia University dataset and the category wise results for the nine classes are summarized in Table~\ref{Pavia}. From the analysis, we observed that the spatial-spectral algorithm using GML gives the best overall accuracy (99.52\%). The variations in the accuracies for different regions depends largely on the area of that particular region. For instance, a comparatively bigger structure belonging to the painted metal sheets class in the center of the image is mostly represented by one region, which leads to a significant improvement on combining it with the segmentation map. On the contrary, regions like the asphalt and bitumen are spread across a large number of regions. 

\begin{table}[!thpb]
\renewcommand{\arraystretch}{1.0}
\caption{Comparison of the overall accuracies of our methods}
\label{results}
\centering
 \begin{tabular}{|c|c|c|} 
 \hline
Classifiers & Indian Pines & Pavia Scene \\[0.5ex]
\hline 
GML             & 97.46\% & 99.03\% \\
GML-W   & 98.31\% & 99.52\% \\ 
CNN             & 87.08\% & 97.39\% \\
CNN-W   & 97.7\% & 99.44\% \\ 
\hline
\end{tabular}
\end{table}

A summary of the overall results obtained on the two datasets are given in Table~\ref{results}. Our experimental results show that the GML based approach outperforms the CNN based approach for both the datasets. This is because the GML performs the classification on a pixel by pixel basis and hence it precisely assigns most of the pixels to their correct classes. However, the CNN technique averages out individual pixel values during the convolution step and so it classifies many of the pixels into neighboring regions which leads to lower accuracies. Our technique also proved to be better than the traditional classifiers of GML and CNN, which gave accuracies of 99.03\% and 97.39\% respectively. As for the Indian Pines dataset, our proposed approach led to a significant improvement of 10.6\% over the traditional CNN based classification approach. In addition to this, our method performed considerably better than the conventional GML classifier with an overall accuracy of 98.31\% on the Indian Pines dataset.

\begin{table}[!thpb]
\renewcommand{\arraystretch}{1.0}
\caption{Comparison with existing deep learning based methods}
\label{results2}
\centering
 \begin{tabular}{|c|c|c|} 
 \hline
Classifiers & Indian Pines & Pavia Scene \\[0.5ex]
\hline 
3D CNN \cite{li2017spectral}             & 99.07\% & 99.39\% \\
CNN-MRF \cite{cao2018hyperspectral}     & 99.27\% & 99.55\% \\
Ours (GML-W)             & 98.31\% & 99.52\% \\
\hline
\end{tabular}
\end{table}

In order to have a more detailed analysis, we compare our approach to the existing deep learning based methods in Table~\ref{results2}. As seen from the empirical results on the Pavia university scene, our GML based spatial-spectral methodology, without any kind of deep learning aspects has an accuracy of 99.52\%, which is better than the performance of 99.39\% achieved by the 3D CNN based spectral classification technique proposed by Li et al.~\cite{li2017spectral}. Our method's performance is also comparable to the state-of-the-art deep learning based technique of CNN-MRF given by Cao et al. \cite{cao2018hyperspectral} that has an accuracy of 99.55\%, even though our approach does not require any sort of expensive training or learning procedures. However, our method achieves an accuracy of 98.31\% on the Indian Pines dataset which is lesser than the accuracy of 99.27\% given by the existing deep learning method because our approach is more suitable for images where the distribution of pixels amongst the different classes is comparatively uniform. 

\section{Conclusion}

In this paper, we present a novel spatial-spectral methodology for the classification of land cover using hyperspectral satellite image data. The proposed approach combines the results of a pixel-wise spectral classification and segmentation maps of the Watershed algorithm, by performing a majority vote on the initial classification output using adaptive neighborhoods defined by the segmentation map. We investigated GML, a machine learning technique as well as CNN, a popular deep learning approach for the pixel-wise image classification, and found that by incorporating the spatial information with the spectral data, the overall classification map contains more homogeneous regions, as compared to only pixel-wise classification of the hyperspectral images. The empirical analyses indicate that our approach performs better than the results discussed in Section 2 for the spatial-spectral classification of hyperspectral images, with an overall accuracy of 99.52\% on the Pavia University dataset and up to 98.31\% on the Indian Pines dataset. Experimental results also demonstrate that our proposed GML based framework - without any kind of expensive training or learning procedures, achieves comparable performance to the state-of-the-art deep learning based methods and is hence a computationally efficient alternative for the classification of hyperspectral image data.

In order to further improve the overall classification, different feature extraction methods need to be investigated so as to find the most effective features for segmentation. Also, our proposed approaches should be verified on multispectral satellite image data to analyze their robustness and in this direction, work is in progress.

\bibliographystyle{splncs04}
\bibliography{sample-base}
%
%

%
%
%
%
\end{document}